\documentclass{article}

 \usepackage[preprint,nonatbib]{nips_2018}

\usepackage[utf8]{inputenc} 
\usepackage[T1]{fontenc}    
\usepackage{hyperref}       
\usepackage{url}            
\usepackage{booktabs}       
\usepackage{amsfonts}       
\usepackage{nicefrac}       
\usepackage{microtype}      

\usepackage[utf8]{inputenc} 
\usepackage[T1]{fontenc}    
\usepackage{hyperref}       
\usepackage{url}            
\usepackage{booktabs}       
\usepackage{amsfonts}       
\usepackage{nicefrac}       
\usepackage{microtype}      

\usepackage{comment} 

\usepackage{amsmath,graphicx}

\usepackage{pstricks,graphicx}
\usepackage{setspace}
\usepackage{psfrag}
\usepackage{amssymb}
\usepackage{amsmath}
\usepackage{amsthm}

\newtheorem{remark}{Remark}

\newtheorem{definition}{Definition}

\title{Effectiveness of Scaled Exponentially-Regularized \\ Linear Units (SERLUs)}

\graphicspath{{figures/}}

  \author{Guoqiang.~Zhang\\
  Center of Audio, Acoustic and Vibration (CAAV)\\
  University of Technology Sydney\\
  Sydney,  Australia \\
  \texttt{guoqiang.zhang@uts.edu.au} \\
   \And
  Haopeng Li \\ Qamcom Research and Technology AB \\
      Isafjordsgatan 20, Kista,  Sweden \\
  \texttt{haopeng.li@qamcom.se} \\
}

\begin{document}

\maketitle

\begin{abstract}
Recently, self-normalizing neural networks (SNNs) have been proposed with the intention to avoid batch or weight normalization. The key step in SNNs is to properly scale the exponential linear unit (referred to as SELU) to inherently incorporate normalization based on central limit theory. SELU is a monotonically increasing function, where it has an approximately constant negative output for large negative input. In this work, we propose a new activation function to break the monotonicity property of SELU while still preserving the self-normalizing property. Differently from SELU, the new function introduces a bump-shaped function in the region of negative input by regularizing a linear function with a scaled exponential function,  which is referred to as a scaled exponentially-regularized linear unit (SERLU). The bump-shaped function has approximately zero response to large negative input while being able to push the output of SERLU towards zero mean statistically. To effectively combat over-fitting, we develop a so-called \emph{shift-dropout} for SERLU, which includes standard dropout as a special case.  Experimental results on MNIST, CIFAR10 and CIFAR100 show that SERLU-based neural networks provide consistently promising results in comparison to other 5 activation functions including ELU, SELU, Swish, Leakly ReLU and ReLU.       
\end{abstract}

\vspace{-3mm}
\section{Introduction}
How to effectively train a deep neural network has been a challenging task.  The training process could be affected by various factors such as the nature of nonlinear activation functions,  weight initialization,  neural network architectures, and optimization methods like stochastic gradient descent (SGD). In the past few years, different techniques have been proposed to improve the training process from different perspectives. Considering selection of the activation function,  the rectified linear unit (ReLU) was found to be much more effective than the binary unit in feed-forward neural networks (FNNs) and convolutional neural networks (CNNs)  \cite{Nair10ReLU}. Careful weight initialization based on the properties of the activation function and layerwise nueron-number has also been found to be essential for fast training (e.g., \cite{He15WeiInti}). Nowadays, neural networks with shortcuts (e.g., ResNet \cite{He15ResNet} and Unet \cite{Ronneberger15Unet}) become increasingly popular as introduction of the shortcuts greatly alleviates the issue of gradient vanishing or explosion, which become severe issues when training extremely deep neural networks.  From the optimization point of view, SGD is simple and effective but often needs manual tunning of its learning rate. Advanced gradient based methods (e.g., Adam \cite{Kingma17}, AdaGrad \cite{Duchi11AdaGrad},  RMSProp \cite{Tieleman12RMSProp}) have thus been proposed to enable both fast training speed and adaptive learning rates.   

In recent years, a family of normalization techniques have been proposed to accelerate the training process. The motivation behind these techniques is to make proper adjustment at each individual layer so that either the input or output statistics of the activation functions of the layer are unified in terms of the first and/or second moments. By doing so, the problem of internal covariance shift can be largely alleviated, thus significantly improving the efficiency of the back-propagation optimization methods. Those techniques can be roughly classified as (a): data-driven normalization, (b): weight normalization, and (c): activation-function normalization. Data-driven normalization operates directly on the  layer-wise internal features of training data, which includes for example batch normalization \cite{Ioffe15BN} and layer normalization \cite{Ba16LN}. This type of normalisations are shown to be remarkably effective but introduce extra computation and often have to carefully handle the inconsistency between training and inference, as the input statistics at the inference stage might be changed due to fewer number of input samples. In \cite{Salimans16WN}, the authors consider performing weight normalization instead to indirectly regulate the statistics of the layer-wise internal features. A mixture of batch and weight normalization has recently been found to work effectively in training large-scale deep neural networks \cite{Hoffer18N}.  
 
The concept of activation-function normalization has been introduced recently by Klambauer at al. \cite{Klambauer17SNN}. The authors proposed a scaled exponential linear unit (SELU), which takes the form of 
\begin{align}
\textrm{SELU}(x) = \lambda_{selu}\left\{\begin{array}{cc} x & x\geq 0 \\ \alpha_{selu}(e^x-1) & \textrm{otherwise} \end{array} \right.,\label{equ:selu_def}
\end{align}
where $\alpha_{selu}\approx 1.6733$ and $\lambda_{selu}\approx 1.0507$. Theoretical analysis is provided showing that if the input to SELU follows a Gaussian distribution with mean and variance floating around 0 and 1,  the mean and variance of the output tend to get closer to 0 and 1 under certain conditions of the weights. The above \emph{self-normalizing} property simultaneously addresses the covariate shift and vanishing or exploding gradient problems across neural layers, thus making data-driven or weight normalizations unnecessary or less important. It is noted that SELU is a monotonically increasing function, and has an approximately constant response $-\lambda_{selu} \alpha_{selu}$ for large negative input. The property of negative constant response is not consistent with that of ReLU which has zero response for negative input. One natural question is if an activation function which has zero response for large negative input while keeping the self-normalizing property would make the resulting neural networks more learnable for training and more generalizable for inference.   

\begin{figure}[t!]
\centering
\includegraphics[width=100mm]{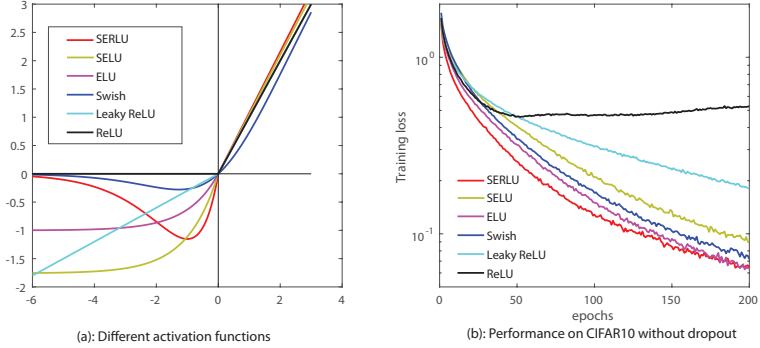}
\caption{ Subplot $(a)$ shows 6 different activation functions where their main differences sit in their responses to negative input. Subplot $(b)$ demonstrates the convergence rates of the 6 activation functions on training data of CIFAR 10 with augmented data. A variant of LeNet architecture is exploited as explained in the experimental section. The subplot suggests that SERLU with the bump-shaped functional segment helps with the training convergence speed.}
\label{fig:AF_con}
\vspace{-5mm}
\end{figure}

In this paper, we develop a new activation function termed as a scaled exponentially regularized linear unit (SERLU). The response of SERLU for negative input is designed to be a linear function regularized by an exponential function, leading to a bump-shaped function for negative input. The bump-shaped function is then properly scaled to be able to push the output of SERLU towards zero mean statistically while having an approximately zero response to large negative input. The self-normalizing property of SERLU is investigated through numerical evaluation to provide a general understanding of its normalizing behavior. The design of SERLU is partly inspired by two recent studies on activation functions \cite{Ramachandran17Swiss} and \cite{Manessi18SAF}. In both works, the authors have attempted to develop new activation functions by either evaluating or learning different combinations of basic functions. Most discovered activation functions do not have monotonicity properties. That is those functions have local bumps, which provide supports for the newly proposed SERLU. We will evaluate the so-called \emph{Swish} activation function discovered in \cite{Ramachandran17Swiss}, which takes the form $\textrm{Swish}(x)=x \cdot {sigmoid}(\beta x)$. In this work, we set $\beta=1$ for  $\textrm{Swish}(x)$. 

Due to the special functional form of SERLU, we propose a new dropout technique for SERLU, referred to as \emph{shift-dropout}. Instead of randomly setting a unit activation to zero as performed by standard dropout, shift-dropout randomly sets a unit activation to the minimum functional value of SERLU, which then undergoes an affine transformation to preserve the mean of unit activations. The above operation attempts to make the output of SERLU at each neuron have a flat distribution rather then concentrate in a small region, making the resulting neural network generalizable. Shift-dropout includes the standard dropout as a special case by setting its minimum functional value to zero.   

In the experiment, we evaluate 6 activation functions including SERLU, SELU\cite{Klambauer17SNN}, ELU \cite{Clevert16ELU}, Swish \cite{Ramachandran17Swiss}, Leakly ReLU \cite{Maas13Leaky}, and ReLU due to their similar responses to positive input (See Figure \ref{fig:AF_con}). We firstly compare SERLU and SELU for a FNN using MNIST due to the fact that both activation functions are derived based on certain properties of FNNs. It is clear from the results that SERLU outperforms SELU.  We then evaluate all the 6 activation functions for a variant of LeNet \cite{LeCun98LeNet} (i.e., one type of CNNs) on MNIST, CIFAR10, and CIFAR100.  The results suggest that SERLU provides consistently promising performance while others do not always produce competitive performance over the three datasets. 

\vspace{-2mm}
 \section{Scaled Exponentially-Regularized Linear Units (SERLUs)}
\vspace{-2mm}   
 \subsection{Expression of SERLU}
 Formally, we define the new activation function as   
\begin{align}
\textrm{SERLU}(x) = \lambda_{serlu}\left\{\begin{array}{cc} x & x\geq 0 \\ \alpha_{serlu} xe^x & \textrm{otherwise} \end{array} \right., \label{equ:serlu_def}
\end{align}
where the two parameters $ \lambda_{serlu}>0$ and $ \alpha_{serlu}>0$ remain to be specified.  Regardless of the two parameters, the difference between  (\ref{equ:selu_def}) and (\ref{equ:serlu_def}) sits in the functional segment for negative input $x< 0$. Specifically, SERLU has a bump-shaped function formulated as $xe^{x}$ for $x< 0$ while SELU has a monotonically increasing function $(e^{x}-1)$ in the same support region. The bump-shaped function ensures that SERLU has a negligible response for large negative input, which is asymptotically consistent with that of ReLU. When $x=-1$, SERLU reaches the global minimum functional value $f_{\min}=-\lambda_{serlu}\alpha_{serlu}e^{-1}$. 

Next we consider the difference between SERLU and $\textrm{Swish}(x)$ as introduced in \cite{Ramachandran17Swiss}. Even though $\textrm{Swish}(x)=x \cdot {sigmoid}(\beta x)$ also has a bump for negative input, its functional form is less flexible than $\textrm{SERLU}(x)$. As will be explained later on, the two parameters $ \lambda_{serlu}>0$ and $ \alpha_{serlu}>0$ makes it possible to freely adjust the mean and variance of the output of $\textrm{SERLU}(x)$ for a random input variable $x$. It is this freedom that allows us to properly  scale $\textrm{SERLU}(x)$ to have the  self-normalizing property. On the other hand, $\textrm{Swish}(x)$ only has one free parameter $\beta$ for adjustment, which is not primarily designed to gain the self-normalizing property.

We note that one can develop alternative bump-shaped functions for negative input to replace $\textrm{SERLU}(x)$.  For instance, one can choose $x^ne^{-x^2}$ or $x^ne^{x}$, where $n$ is an odd number. Another bump-shaped functional form would be to combine a shift version of $-e^x$ and a linear functional segment. We conjecture that the above functional forms might produce similar promising experimental results as $\textrm{SERLU}$ as will be demonstrated later on. In this work, we choose the form $xe^{x}$ due to its simplicity. Furthermore, $xe^{x}$ facilitates computation of the mean and variance of the output of SERLU for a Gaussian random input variable $x$.  

\vspace{-1mm}
\subsection{Input-output statistical properties of SERLU}
\vspace{-1mm}

 We focus on the FNNs where the weights are not shared across space to facilitate the analysis. We follow a similar procedure as \cite{Klambauer17SNN} for analyzing SELU. The basic idea of \cite{Klambauer17SNN} is to assume both the input $x$ and output $z=f(x)$ for any nueron in a neural network are random variables.  Furthermore, the output random variables $\{z_i\}$ from the same neural layer are to assume to be independent. If the number of nuerons at every layer is large enough, the input random variable $x$ to a nueron in a certain layer can be assumed to approximately follow a Gaussian distribution as it is a linear combination of many random variables from the layer below. 
Since there are often many neurons per layer in a FNN,  the Gaussian distribution approximation is reasonable.    

We now consider two consecutive layers where the connections are characterized by a weight matrix $\boldsymbol{W}$.  Suppose the lower layer has $n$ units with random output variables $\{z_{i,low}| 1\leq i\leq n \}$, where we use $\boldsymbol{z}_{low}$ to denote its compact vector form. Following the work \cite{Klambauer17SNN}, we assume all the output variables from the lower layer are independent, and have the same mean $\mu=\textrm{E}(z_{i,low})$ and the same variance $\nu = \textrm{Var}(z_{i,low})$, where the notations $\textrm{E}(\cdot)$ and $\textrm{Var}(\cdot)$ represent expectation and variance of a random variable, respectively. Given $\boldsymbol{z}_{low}$ from the lower layer, the input $x_{up}$ and output $z_{up}$ for a neuron at the upper layer can be represented as 
\begin{align}
z_{up}=\textrm{SERLU}(x_{up}),\quad \textrm{ where } x_{up} =\boldsymbol{z}_{low}^T\boldsymbol{w}= \sum_{i=1}^n w_i z_{i,low}, \label{equ:x_y_relation}
\end{align}
where $\boldsymbol{w}$ is one column vector of  $\boldsymbol{W}$.  Similarly to $z_{i,low}$, we denote the mean and variance of the output variable $z_{up}$ as $\tilde{\mu}$ and $\tilde{\nu}$. To facilitate the analysis later on, we let $\omega =1^T\boldsymbol{w}= \sum_{i=1}^n w_i$  and  $\tau  = \boldsymbol{w}^T\boldsymbol{w}= \sum_{i=1}^n w_i^2$. Most of the above introduced notations are in line with those in \cite{Klambauer17SNN} to enhance readability of the paper.   
 
Finally we would like to derive the expressions for $(\tilde{\mu}, \tilde{\nu})$ for the output $z_{up}$ based on the statistics of $\boldsymbol{z}_{low}$.  To do so, we first consider the input variable $x_{up}$. By using the independence assumption of $z_{i,low}$, the mean and variance of $x_{up}$ are expressed as $\textrm{E}(x_{up})=\mu\omega$ and $\textrm{Var}(x_{up})=\nu\tau$. Similarly to \cite{Klambauer17SNN}, we assume $x_{up}$ has a Gaussian distribution (i.e., $x_{up}\sim \mathcal{N}(\mu\omega, \sqrt{\nu\tau})$). The central limit theorem (CLT) suggests that a large number of neuorns in the lower layer leads to an accurate Gaussian approximation. In FNNs, it is common to have hundreds of nuerons per layer. Accordingly, the expressions for $(\tilde{\mu}, \tilde{\nu})$ for the output $z_{up}$ can be computed as
\begin{align}
\tilde{\mu} &=\hspace{-1mm} \int_{-\infty}^{0} [\lambda_{serlu}\alpha_{serlu} xe^x] f_{Gauss}(x,\mu\omega, \sqrt{\nu\tau})dx \hspace{-1mm}+\hspace{-1mm} \int_{0}^{\infty} [\lambda_{serlu} x] f_{Gauss}(x,\mu\omega, \sqrt{\nu\tau})dx  \nonumber  \\
\tilde{\nu} &= \hspace{-1mm}\int_{-\infty}^{0} [\lambda_{serlu}\alpha_{serlu} xe^x]^2 f_{Gauss}(x,\mu\omega, \sqrt{\nu\tau})dx \hspace{-1mm}+\hspace{-1mm} \int_{0}^{\infty} [\lambda_{serlu} x]^2 f_{Gauss}(x,\mu\omega, \sqrt{\nu\tau})dx -\tilde{\mu}^2. \nonumber 
\end{align}
By using algebra, the expression $\tilde{\mu}$ and $\tilde{\nu}$ can be simplified as
\begin{align}
&\hspace{0mm}\tilde{\mu} =\frac{\lambda_{serlu}}{2\sqrt{\pi}}e^{-\frac{\mu ^2 \omega ^2}{2 \nu  \tau }}  \Bigg(e^{\frac{(\nu  \tau +\mu  \omega )^2}{2 \nu  \tau }} \sqrt{\pi } \alpha_{serlu}
 (\nu  \tau +\mu\omega)\text{Erfc}\left[\frac{\nu
 \tau +\mu  \omega }{\sqrt{2} \sqrt{\nu  \tau }}\right] +\sqrt{2} \sqrt{\nu  \tau } (1-\alpha_{serlu}) \nonumber \\
&\hspace{46mm}+e^{\frac{\mu ^2 \omega ^2}{2 \nu  \tau }} \sqrt{\pi } \mu  \omega  \left(2-\text{Erfc}\left[\frac{\mu\omega }{\sqrt{2} \sqrt{\nu  \tau }}\right] \right)\Bigg), \label{equ:mu_map} \\
&\hspace{0mm}\tilde{\nu}  = \hspace{-1mm}\frac{1}{2} \lambda_{serlu} ^2 \Bigg(\hspace{-1mm} e^{-\frac{\mu ^2 \omega ^2}{2 \nu  \tau }} \sqrt{\frac{2}{\pi }} \sqrt{\nu  \tau } [(1\hspace{-0.6mm}-\hspace{-0.6mm} \alpha_{serlu} ^2)\mu  \omega \hspace{-0.6mm} -\hspace{-0.6mm}  2\alpha_{serlu} ^2 \nu  \tau ]  
\hspace{-0.6mm}+\hspace{-0.6mm} \left(\nu\tau +\mu ^2 \omega ^2\right)\left(2\hspace{-0.6mm}-\hspace{-0.6mm}\text{Erfc}\left[\frac{\mu  \omega }{\sqrt{2} \sqrt{\nu  \tau }}\right]\right) \nonumber \\
&\hspace{5mm}+  \alpha_{serlu} ^2  e^{-\frac{\mu^2 \omega ^2}{2 \nu  \tau }}e^{\frac{(2 \nu  \tau +\mu  \omega
)^2}{2 \nu  \tau }} \left(4 \nu ^2 \tau ^2+\mu ^2 \omega ^2+\nu  (\tau +4 \mu  \tau  \omega )\right) \text{Erfc}\left[\frac{2 \nu  \tau
+\mu  \omega }{\sqrt{2} \sqrt{\nu  \tau }}\right]\Bigg)- \tilde{\mu}^2. \label{equ:nu_map}
\end{align}
It is clear that $(\tilde{\mu} ,\tilde{\nu})$ is a function of $({\mu},{\nu})$, $({\omega},{\tau})$ and $(\alpha_{serlu},\lambda_{serlu})$. 

\subsection{Determination of $\alpha_{serlu}$ and $\lambda_{serlu}$}
So far we have characterized the relationship between $(\tilde{\mu} ,\tilde{\nu})$ for the upper layer and  $({\mu} ,{\nu})$ for the lower layer.  To make SERLU exhibit the self-normalizing property as SELU, the first step would be to determine $\alpha_{serlu}$ and $\lambda_{serlu}$ so that the mean and variance of the output of SERLU are consistent across layers, i.e., $(\tilde{\mu} ,\tilde{\nu}) =  ({\mu} ,{\nu})$.  In other words, $\alpha_{serlu}$ and $\lambda_{serlu}$ should be selected to ensure that $({\mu} ,{\nu})$ is a fixed point of the mapping $g: ({\mu} ,{\nu})\rightarrow (\tilde{\mu} ,\tilde{\nu})$ as defined by (\ref{equ:mu_map})-(\ref{equ:nu_map}). In principle, one can specify different fixed points and then search for the corresponding solution $\alpha_{serlu}$ and $\lambda_{serlu}$. In this work, we set the mean $\mu=0$ and the variance ${\nu} =1$ for the fixed point, which is consistent with derivation of SELU. Furthermore, we assume $\boldsymbol{w}$ is a normalized weighting vector with $\omega = 0$ and $\tau = 1$. With the above parameter specification, (\ref{equ:mu_map})-(\ref{equ:nu_map}) becomes two equations w.r.t. $(\alpha_{serlu} ,\lambda_{serlu})$. Solving them produces 
\begin{align}
\alpha_{serlu} \approx 2.90427 \quad\textrm{ and } \quad \lambda_{serlu} \approx 1.07862. \label{equ:alpha_lambda}
\end{align}
Algebraically speaking, the parameter $\alpha_{serlu}$ regulates the mean  unit activations to be zero while $\lambda_{serlu}$ normalizes the variance of the unit activation. 

Next we study if $(\mu, \nu) =(0,1)$ is a stable fixed point under the mapping $g$. It is know that if the spectral norm of the Jacobian of the mapping $g$ is less than 1, then the fixed point is stable and $g$ is a contraction mapping in a small region around the fixed point. By computation, the $2\times 2$ Jacobian $\mathcal{J}(\mu,\nu)$ at the fixed point $(0,1)$ with the parameter setup (\ref{equ:alpha_lambda}) is $\mathcal{J}(0,1)=((0,0.194557),(0, 0.605258))$. The spectral norm of $\mathcal{J}(0,1)$ is thus 0.635758 <1. This implies that $g$ is indeed a contraction mapping around $(0,1)$. It is reported in \cite{Klambauer17SNN} that the spectral norm of $\mathcal{J}(0,1)$ for SELU is 0.7877 (>0.635758). This suggests that SERLU has comparable convergence as SELU in the region close to $(0,1)$. In general, the convergence rate is also affected by $(\omega,\tau)$. 

We notice that batch normalization explicitly operates on the layerwise internal features of training data. As a result, backpropogation of gradient needs to take into account the operation, which introduce extra computational burden. On the other hand, both SERLU and SELU implicitly absorb the normalization operation into their functional form via two fixed scalar parameters, which is more computationally efficient.      

\vspace{-1mm}
\subsection{On self-normalizing property of SERLU through numerical evaluation}
\vspace{-1mm}

In this subsection, we consider the cross-layer effect of SERLU in a FNN. Firstly,  we need to relax the assumption on a normalized weight vector $\boldsymbol{w}$. Specifically, we assume the mean $\omega$  and second moment $\tau$ of the vector $\boldsymbol{w}$ may fluctuate around  $0$ and $1$, respectively. In practice, SGD based optimization methods deliberately introduce noise to the gradient of $\boldsymbol{w}$ via minibatches to  alleviate over-fitting, leading to disturbance of $\omega$ and $\tau$ per iteration. We are interested in if $g: ({\mu} ,{\nu})\rightarrow (\tilde{\mu} ,\tilde{\nu})$ defined in (\ref{equ:mu_map})-(\ref{equ:nu_map}) is still an attracting mapping under the disturbance. If so, the resulting SERLU-based FNN would preserve normalization across layers, leading to a SNN.   

We first review the definition of SNNs introduced in \cite{Klambauer17SNN}:
\begin{definition}[Self-normalizing neural networks]
A neural network is self-normalizing if it possesses a mapping $g: \Omega\rightarrow \Omega$ for each activation $z$ that maps mean and variance from one layer to the next and has a stable and attracting fixed point depending on $(\omega,\tau)$ in $\Phi$, where $\Phi=\{(\omega, \tau) | \omega\in [\omega_{\min},\omega_{\max}], \tau \in [\nu_{\min},\tau_{\max}] )\}$. Furthermore, the mean and variance remain in the domain $\Omega$, that is $g(\Omega)\subset \Omega$, where $\Omega = \{(\mu, \nu) | \mu\in [\mu_{\min},\mu_{\max}], \nu \in [\nu_{\min},\nu_{\max}] )\}$.
\end{definition}

The above definition implies that if the disturbance to the weight vectors are uniformly bounded (w.r.t.  the mean $\mu$ and second moment $\tau$) across neural layers, the mean and variance of the unit activations stay in the predefined intervals when propagating them from the bottom layer towards the top one in the network. Furthermore, if the mean and second moments of the weight vectors converge over iterations, the mean and variance of the unit activations at every layer converge to their respective fixed points. In practice, weight normalization is one solution to make the weight vectors free of disturbance over iterations. Differently from batch normalization,  weight normalization only requires a small amount of computational resource. 

We now study the properties of the mapping $g$ for SERLU by numerical evaluation. We let the two domains $\Omega$ and $\Phi$ to be  $\Omega = \{ \mu\in [\mu_{\min},\mu_{\max}]=[-0.2,0.2], \nu \in [\nu_{\min}, \nu_{\max}]=[0.8, 1.5] \}$ and $\Phi = \{\omega\in [\omega_{\min}, \omega_{\max}]= [-0.1, 0.1], \tau\in [\tau_{\min,\tau_{\max}}]=[0.9,1.2]\}$. Our primary interest is the maximum spectral norm of the Jacabian of $g$ and the domain $g(\Omega)$ after mapping.  To achieve the goal, we first generate a dense sampling grid of  $(\mu,\omega,\nu,\tau)$ in $\Omega$ and $\Phi$ and then numerically evaluate the output $(\tilde{\mu},\tilde{\nu})$ for each sampling point from the grid.  The sampling interval is set as 0.02 for each variable.  The results are summarized in Table~\ref{table:SN_SERLU}.  It is found that the maximum spectral norm over all the grid points is 0.7837<1, implying that the mapping $g$ is attractive around each evaluated point. In addition, the bounds for $\tilde{\mu}$ and $\tilde{\nu}$ are within the domain $\Omega$ for all the tested grid points. The above results suggest that the response of the mapping $g$ at those grid points is controllable.

\begin{table}[h]
\caption{  Numerical evaluation of the mapping $g$ for SERLU. Both the maximum spectral norm and the bounds of $\tilde{\mu}$ and $\tilde{\nu}$ are obtained at the boundary points of the two domains $\Omega$ and $\Phi$.  } 
\label{tab:time}
\centering
\begin{tabular}{|c|l|}
\hline
\textrm{Sampling grid }
&\hspace{-3.5mm} {{ $\begin{array}{cc}\mu=\{-0.2:0.02:0.2\}, &  \tau=\{0.9:0.02:1.2\}  \\  \omega=\{-0.1:0.02:0.1\}, &  \nu=\{0.8:0.02:1.5\}  \end{array}$}}\hspace{-3mm}  \\
\hline  
\hspace{-1.5mm}{Max. spectral norm of $\mathcal{J}(\mu,\nu)$ } \hspace{-1.5mm} & \hspace{5mm}{0.7837} $(\mu=-0.2,\;\omega=-0.1,\;\nu=0.8,\;\tau=1.2)$  \\ 
\hline
\hspace{-1.5mm}{lower bound of $\tilde{\mu}$} &\hspace{1mm} {$ -0.0751$  $(\mu=\hspace{3mm}0.2,\;\omega=-0.1,\;\nu=0.8,\;\tau=0.9)$}  \\
\hline
\hspace{-1.5mm}{upper bound of $\tilde{\mu}$} & {\hspace{4.5mm}$ 0.1629$ $(\mu=\hspace{3mm} 0.2,\;\omega=\hspace{3mm}0.1,\;\nu=1.5,\;\tau=1.2)$ }  \\
\hline
\hspace{-1.5mm}{lower bound of $\tilde{\nu}$} &\hspace{3.5mm} {$0.8125 $ $(\mu=\hspace{3mm}0.2,\;\omega=-0.1,\;\nu=0.8,\;\tau=0.9)$}  \\ 
\hline
\hspace{-1.5mm}{upper bound of $\tilde{\nu}$} & \hspace{4.5mm}{$1.4551$ $(\mu=\hspace{3mm}0.2,\;\omega=\hspace{3mm}0.1,\;\nu=1.5,\;\tau=1.2)$}  \\ 
\hline
\end{tabular}
\label{table:SN_SERLU}
\vspace{0mm}
\end{table}      

\begin{remark}
We point out that the numerical evaluation is mainly for drawing a general picture of the response of the mapping function $g$ over the considered domains $\Omega$ and $\Phi$.  It is not a proof for showing that $g$ is an attractive mapping.  We conjecture that a similar theoretical analysis as that of SELU exists for SERLU. This is because the two activation functions only involve three basic functions $x$, $e^x$ and $xe^{x}$, which can be nicely combined with the functional form $e^{-x^2/2}$ of Gaussian distribution in computing the mean and variance of two unit activations. 

\end{remark}

 
 \vspace{-2mm}
\section{Specialized Dropout Technique for SERLU} 
\vspace{-1mm}
\label{sec:betaDropout}

It is well known that dropout plays an important role to avoid or alleviate the overfitting problem.  Since the pioneering work by Srivastava et al. \cite{Srivastava14Dropout}, the standard dropout operation has been modified or adjusted in different ways for training special types of neural networks. One typical example is to redesign dropout for training LSTM-based neural networks where the recurrent connection makes training considerably challenging (e.g., \cite{Gal16Dropout,Krueger17ZoneOut}).  In \cite{Klambauer17SNN}, the authors propose a so-called alphaDropout for SNNs to improve the performance, which we will briefly discuss later on.  

In this work, we propose a variant of the standard dropout for SERLU-based neural networks. Let us first revisit the standard dropout. The basic idea is to randomly set an activation $z$ to zero during training with probability $1-q$ where $0<q\leq 1$. To compensate for the dropout effect, the activation is then scaled by $\frac{1}{q}$. By doing so, the mean of the activations would be preserved, ensuring consistency over iterations. The above statement can be easily justified mathematically. Suppose the activation $z$ has mean $\textrm{E}(z)=\mu$, and the dropout variable $d$ follows a binomial distribution $B(1,q)$. It is immediate that the mean $\textrm{E}(d*z/q)=\mu$.  The dropout operation allows to train an ensemble of all sub-networks seamlessly and effectively.

Considering $\textrm{SERLU}(x)$, its minimum functional value $f_{\min}=-\lambda_{serlu}\alpha_{serlu}e^{-1}$ is achieved only at a single point $x=-1$. This suggests that the minimum value would not be frequently visited during training. As a result, it might be a bit difficult for points at the two sides of $x=-1$ jump to the opposite side to fully explore the functional properties of SERLU in the whole negative input region $x<0$. Based on the above analysis, we extend the standard dropout in a following manner. Instead of zero, we randomly set an activation $z$ to the minimum functional value $f_{\min}$ with probability $1-q$.  We use $\tilde{z}$ to denote the modified activation. The mean of $\tilde{z}$ is thus given by $\textrm{E}(\tilde{z})=\textrm{E}(d*z + (1-d)*f_{\min}) = q\mu+(1-q)f_{\min} $, which is not equal to $\mu$. To be able to preserve the mean $\mu$ of the activation, one simple solution is to perform an affine transformation to $\tilde{z}$, which can be mathematically expressed as 
\begin{align}
\hat{z} = \frac{1}{q}[ \tilde{z}-(1-q)f_{\min} ].
\label{equ:shiftDropout}
\end{align}
It is not  difficult to show that the final activation $\hat{z}$ indeed preserves the mean $\mu$ as the original activation $z$. We refer to the new dropout technique as \emph{shift-dropout}. 

We point out that alphaDropout in \cite{Klambauer17SNN} also randomly sets an activation $z$ to the minimum functional value of SELU.  One major difference from shift-dropout is that alphaDropout adjusts the activation to follow a predefined mean $\mu=0$ and variance $\nu=1$ to be in line with the normalization properties of SELU. We think that in practice, the input-output statistics of either SELU or SERLU may break the assumption of predefined mean $\mu=0$ and variance $\nu=1$ due to, for example, rough Gaussian distribution approximation or the statistical bias introduced by minibatches.  
Shift-dropout is motivated to preserve the mean $\textrm{E}(z)=\mu$ of the real training data, which may bring robustness when training a SERLU-based neural network. 
  
 \begin{remark} 
From a high level perspective, shift-dropout includes standard dropout as a special case by setting $f_{\min} =0$ in (\ref{equ:shiftDropout}). Application of standard dropout to ReLU and Sigmoid can be interpreted as performing shift-dropout to ReLU and Sigmoid as their minimum functional value is zero. 
\end{remark}

\vspace{-1mm}
\section{Experimental Results}
\vspace{-1mm}

\begin{figure}[t!]
\centering
\includegraphics[width=130mm]{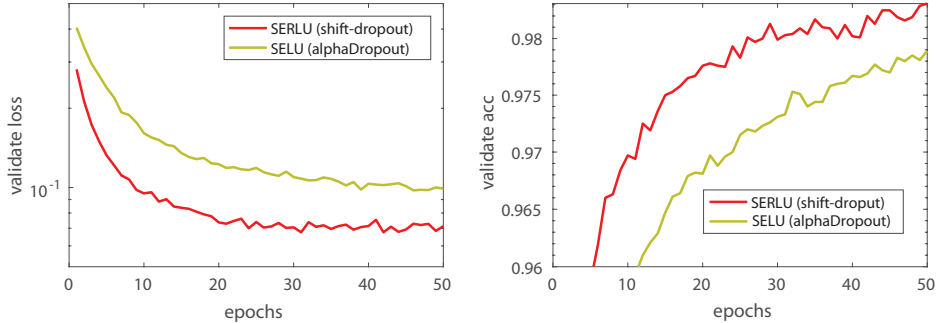}
\caption{Performance of SERLU and SELU for a FNN over MNIST.}
\label{fig:FNN_compare}
\vspace{-2mm}
\end{figure}

\textbf{Evaluation of SERLU and SELU over a FNN:} In the first experiment, we investigated the performance of the two functions over a FNN of 4 hidden layers with 200 neurons per layer  for the handwritten digital classification task from MNIST.  We motivation is that since both SERLU and SELU are designed based on the assumption of many neurons per layer to allow for the application of the central limit theory, it is interesting to find out their performance on a FNN. In the implementation, dropout with 10\% rate was introduced at every hidden layer. SELU used alphaDropout while SERLU used shift-dropout to fully explore their capability.  The network was trained using RMSProp \cite{Tieleman12RMSProp} with a learning rate of $10^{-4}$ and  a decay of $10^{-6}$.  Figure \ref{fig:FNN_compare} displays their convergence rates for the validation dataset. It is clear that SERLU exhibits faster convergence speed, which might be due to the effect of the  bump-shaped functional segment of SERLU for negative input.   

\textbf{Evalution of 6 activation functions over a CNN:} In the 2nd experiment, we evaluated 6 activation functions: namely SERLU, SELU, ELU, Swish, Leaky-ReLU, and ReLU. All the six functions have similar responses to positive input while their responses to negative input are designed by following different methodologies (see Figure \ref{fig:AF_con}:(a) for their respective response curves).  It is therefore of great interest to evaluate their performance within a unified framework.   
 
We consider the classification problem over three datasets: MNIST,  CIFAR10 and CIFAR100. The tested neural network for CIFAR10 and CIFAR100 is a variant of LeNet \cite{LeCun98LeNet} as illustrated in Table~\ref{table:LeNet} while the network for MNIST is obtained by removing Layer 1 and Layer 3 in the table.  Ideally, different datasets should employ specialized neural network architectures to optimally capture the input-output relationship of each dataset. In our experiment, the architecture of the tested CNN is not optimized to promote the classification accuracy for the three datasets but rather to provide a unified framework to investigate the behaviours of the 6 activation functions. 

Similarly to the first experiment, we trained the network using RMSProp \cite{Tieleman12RMSProp} with a learning rate of $10^{-4}$ and  a decay of $10^{-6}$. To alleviate overfitting, each of the three datasets was augmented with additional training data (e.g., by shifting images vertically and/or horizontally).  SERLU used shift-dropout as proposed in Section \ref{sec:betaDropout} while all other activation functions used the standard dropout  \footnote{\small alphaDropout was found to work only for small dropout rate for SELU, which is consistent with the findings in  \cite{Klambauer17SNN}. }. The dropout rate for each layer is the same across all the 6 activation functions. 

\begin{figure}[t!]
\centering
\includegraphics[width=130mm]{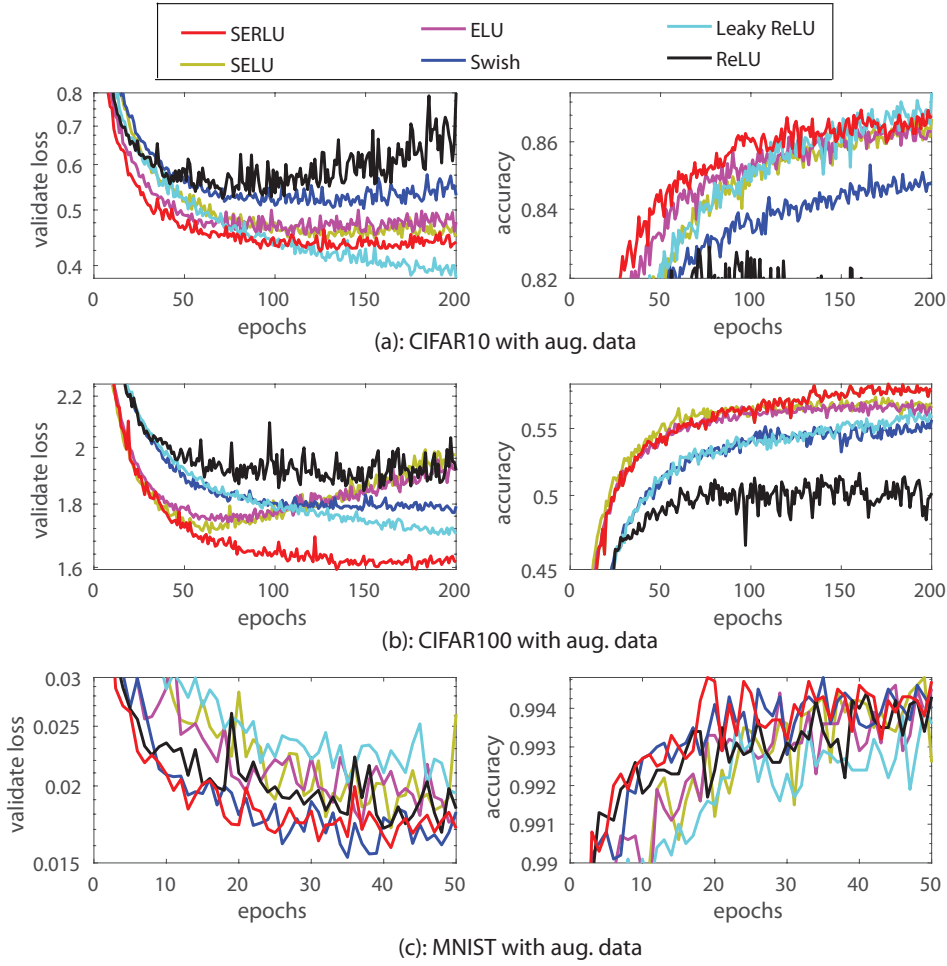}
\caption{Performance of the 6 activation functions over MNIST, CIFAR10 and CIFAR100.}
\label{fig:converge}
\vspace{-2mm}
\end{figure}

\begin{table}[t]
\caption{  A variant of LeNet for evaluating the 6 activation functions over CIFAR10 and CIFAR100} 
\label{tab:time}
\centering
\begin{tabular}{|c|c||c|c||c|c|}
\hline
\hspace{-1.5mm}\textrm{Layer 1} \hspace{-2mm}& \hspace{-2mm} conv.:  $3\times3@ 32$ \hspace{-2mm}&\hspace{-2mm} \textrm{Layer 3} \hspace{-2mm}&\hspace{-3mm} conv.: $3\times3@ 64$  \hspace{-3mm}&\hspace{-3mm}  \textrm{Layer 5} &\hspace{-3mm} $\begin{array}{c}\textrm{dense: 512 neurons} \\\textrm{dropout} \end{array}$ \hspace{-2mm} \\
\hline
\hspace{-2.5mm} Layer 2 \hspace{-2mm}&\hspace{-2mm}$\begin{array}{ccc} \textrm{ conv.: }  3\times3@ 32  \\ \textrm{max-pooling} \\ \textrm{dropout}\end{array}$ \hspace{-2mm}&\hspace{-2mm}  
Layer 4 &\hspace{-3mm}  $\begin{array}{ccc}\textrm{conv.: } 3\times3@ 64  \\ \textrm{max-pooling} \\ \textrm{dropout}\end{array}$  \hspace{-2mm}&\hspace{-2mm}
Layer 6 &\hspace{-2mm} dense + softmax  \\
\hline
\end{tabular}
\label{table:LeNet}
\vspace{-3mm}
\end{table} 

The performance results w.r.t. validation data are demonstrated in Figure \ref{fig:converge}. it is seen that SERLU provides consistently promising performance for the three datasets. In all the three experiments, no noticable overfitting is observed over the tested number of epochs for the new activation function. Furthermore, SERLU provides the best performance for CIFAR100 and the second-best performance for CIFAR10, and performs equally well as Swish for MNIST w.r.t. validation loss. Even though Leaky ReLU converges the fastest for CIFAR10, its performance for the other two datasets is not competitive. SELU and ELU exhibit similar convergence behaviours, which is reasonable given that they have similar activation responses to input. The two activation functions exhibit over-fitting for CIFAR100 which might be because the LeNet is not suitable for CIFAR100, which explains why the overall performance of the 6 activation functions is not promising. 

Finally we note that Swish demonstrates roughly the same fast convergence speed as SERLU for MINST w.r.t. validation loss.  We recall that Swish has a small flat bump for negative input while SERLU has a relatively large sharp bump. The different performance of the two activation functions for CIFAR10 and CIFAR100 suggests that a large bump might be generally favourable when constructing a neural network.   

\section{Conclusions}

In this paper we have introduced a new activation function termed as SERLU together with a new dropout technique termed as shift-dropout. SERLU is designed to have a large sharp bump in the negative input region which pushes the mean of unit activation to zero. This is fundamentally different from ELU and SELU which are monotonically increasing functions, and have approximately constant negative response for large negative input.  Numerical evaluations on a dense sampling grid suggest that SERLU exhibits the self normalising properties as SELU at those tested points. Further theoretical analysis is required to show that SERLU has the self normalising properties over a continuous region. 

Experiments on MNIST, CIFAR10 and CIFAR100 were conducted showing that SERLU has consistently promising results over the three datasets in comparison to other 5 activation functions (ELU, SELU, Swish, Leaky ReLU and ReLU). We believe that the bump-shaped functional segment of SERLU together with the coupled shift-dropout plays a key role in its promising performance. 

As mentioned in Subsection 2.1, there are different ways to introduce bumps to activation functions.  We hope our work could bring insights for interested researchers to design more effective bump-shaped activation functions to benefit the whole deep learning community.   


\end{document}